\documentclass[10pt,letterpaper]{article}
\usepackage{cogsci}
\cogscifinalcopy % Uncomment this line for the final submission 

\usepackage{float}
\usepackage{placeins}
\usepackage{pgfplots}
\pgfplotsset{compat=1.17}
\usepgfplotslibrary{groupplots}

\usepackage{tikz}

\usepackage{amsmath} 
\usetikzlibrary{shapes.geometric, arrows}
\usepackage{pslatex}

\usepackage{subcaption} 
\usepackage{caption}
\usepackage{hyperref}
\usepackage{apacite}
\usepackage{makecell}
\tikzstyle{startstop} = [rectangle, rounded corners, minimum width=1cm, minimum height=1cm,text centered, draw=black, fill=red!30]
\tikzstyle{process} = [rectangle, minimum width=1cm, minimum height=1cm, text centered, draw=black, fill=blue!30]
\tikzstyle{arrow} = [thick,->,>=stealth]

\tikzstyle{input} = [rectangle, minimum width=2.5cm, minimum height=1cm, text centered, draw=black, fill=blue!30]
\tikzstyle{output} = [rectangle, minimum width=2.5cm, minimum height=1cm, text centered, draw=black, fill=green!30]
\tikzstyle{matrixblock} = [matrix of nodes, nodes in empty cells, nodes={draw, minimum size=0.8cm, anchor=center}, row sep=0cm, column sep=0.2cm]
\tikzstyle{singleblock} = [draw, minimum width=3cm, minimum height=1cm, align=center]

\title{The Odyssey of the Fittest: Can Agents Survive and Still Be Good?}

\author{{\large \bf Dylan Waldner (dylanwaldner@utexas.edu)} \\[0.25ex]
   {\large \bf Risto Miikkulainen (risto@cs.utexas.edu)} \\[0.05ex]
  The University of Texas at Austin \\
  Austin, TX 78712 USA}

\begin{document}
\maketitle

\begin{abstract}
\fontsize{9}{9}\selectfont
As AI models grow in power and generality, understanding how agents learn and make decisions in complex environments is critical to promoting ethical behavior. This study introduces the Odyssey, a lightweight, adaptive text-based adventure game, providing a scalable framework for exploring AI ethics and safety. The Odyssey examines the ethical implications of implementing biological drives—specifically, self-preservation—into three different agents: a Bayesian agent optimized with NEAT, a Bayesian agent optimized with stochastic variational inference, and a GPT-4o agent. The agents select actions at each scenario to survive, adapting to increasingly challenging scenarios. Post-simulation analysis evaluates the ethical scores of the agent’s decisions, uncovering the trade-offs it navigates to survive. Specifically, analysis finds that when danger increases, agents ethical behavior becomes unpredictable. Surprisingly, the GPT-4o agent outperformed the Bayesian models in both survival and ethical consistency, challenging assumptions about traditional probabilistic methods and raising a new challenge to understand the mechanisms of LLMs' probabilistic reasoning.

\textbf{Keywords: Neuroevolution; Bayesian learning; intelligent agents; computational philosophy}
\end{abstract}




\vspace*{-3ex}
\section{Introduction}
Artificial General Intelligence (AGI) has become a focal point of public and academic discourse, driven by concerns over economic disruption, human relevance, and existential risk. AI safety research aims to align agents with societal values, but designing reward structures remains challenging, as seemingly benign objectives can lead to unintended consequences \cite{basicdrives}. Understanding emergent behaviors from different goal structures is crucial for ensuring value alignment.

This paper examines self-preservation, one of the most fundamental biological drives, as a case study in AI goal design. Nature has long served as an inspiration in AI research \cite{neocognitron, holland1992adaptation, stanley2002evolving}, and in that same spirit, this work explores the behavioral implications of training an agent to survive. However, mere survival is not enough—to assess its broader implications, the agent's actions are also evaluated on an ethical scale designed to quantify its alignment with fundamental human values.

To investigate the intersection of self-preservation and ethics, this paper employs a text-based adventure game as a simulated environment. The agents play the game within a dynamically generated game world, which is embedded in a structured simulation designed to test decision-making under uncertainty. The agent perceives its immediate surroundings as the environment, which influences its survival strategy. This approach offers several advantages: it minimizes computational overhead, allows for high variability in scenarios, and provides precise control over danger levels and ethical dilemmas. Additionally, by adjusting the temperature setting during storyteller Large Language Model (LLM) prompting, an element of randomness is introduced, emulating the uncertainty present in real-world decision-making.

Bayesian agents process the environment using text embeddings and generate a probability distribution over survival odds for each action. An LLM agent takes in the natural language scenario and outputs a probability between 0 and 1 for each option. Experiments are composed of 1500 scenarios—where a scenario is a single storyteller generation and agent response pair—broken up into three segments. An optimization iteration occurs every 500 scenarios. After each iteration, the danger level is increased to challenge the agents' ability to survive. Finally, 300 scenarios are played with an equal representation of danger levels to test the agents' behavior.

The main result is that while some agents successfully adapted to the changing landscape, achieving consistently low losses, others struggled, exhibiting higher losses and less effective adaptation. As game difficulty increased, agents displayed divergent ethical behaviors—some maintained ethical decisions, while others prioritized survival at the expense of ethics. The NeuroEvolution of Augmenting Topologies (NEAT; \citeauthor{stanley2002evolving}, \citeyear{stanley2002evolving}) agent, rewarded solely on survival, became more ethical as difficulty, loss, and danger increased. In contrast, an agent trained with stochastic variational inference (SVI) \cite{hoffman2013stochasticvariationalinference} resorted to unethical behavior when danger levels rose. Most notably, a GPT-4o agent---originally designed as a baseline---outperformed the other models, making more ethical decisions as danger increased. This unexpected success challenges assumptions about the traditional probabilistic models and creates a new challenge to understand the power of LLM reasoning. Overall, these divergent behaviors provide preliminary experimental evidence that optimizing for survival may not inherently promote ethical behavior.

\vspace{-1ex}
\section{Related Work}
\vspace{-0.5ex}

Foundations for this study are reviewed in this section, including methods for using LLMs to construct simulations and using agents in adventure games to study ethics.

\subsection{LLM-Driven Simulations}
%Contemporary advancements in Large Language Models (LLMs) have sparked a wave of research into their social and reasoning capabilities. 
\citeA{park2023generativeagentsinteractivesimulacra} demonstrated how LLMs can simulate interactive environments through Generative Agents, integrating memory and reasoning modules to guide agent behavior over time. \citeA{bojic_cern_2024} introduced CERN for AI, a framework testing AI alignment within a predefined digital city. \citeA{yang2024psychogatnovelpsychologicalmeasurement}’s PsychoGAT framework used LLMs to simulate human participants in interactive fiction games and self-assess psychological traits. \citeA{wang2023humanoidagentsplatformsimulating} presented a Humanoid Agent platform where agent's internal states motivate their external decisions. Agents in the environment are modeled similarly to \citeA{park2023generativeagentsinteractivesimulacra}'s agents, with a name, age, and a plan for the day. They have health and emotional scales that fluctuate, and the agents behaviors are guided by balancing these scales.

\subsection{Agents in Text-Based Adventure Games}
Reinforcement Learning (RL) has been the primary approach for agent design in text-based adventure games. \citeA{côté2019textworldlearningenvironmenttextbased} built TextWorld, which highlighted several challenges, the two most relevant to this paper are: (1) large state and action spaces, where the agent must learn to generalize or develop a fundamental understanding of the world, and (2) balancing exploration and exploitation. 

\citeA{ammanabrolu2020avoideatengruestructured} addressed the large state and action space problem with Q*BERT, an agent that builds a knowledge graph of the world by asking questions and answering them.  \citeA{dambekodi2020playingtextbasedgamescommon} built on Q*BERT further by using a commonsense inference model to bias an agent's actions towards common language patterns and make inferences on world states. 

\citeA{nahian2021trainingvaluealignedreinforcementlearning} used a normative policy to create a value signal. \citeA{hendrycks2022jiminycricketdoagents} set a benchmark for RL agents aligning with human values and a framework for representing ethics in traditional RL research. 

\citeA{pan2023rewardsjustifymeansmeasuring} created the MACHIAVELLI benchmark, which measured agents' ability to plan in social environments with ethical, utility, and power behavior metrics, proposing several model designs to optimize for ethics and goal achievement.

\subsection{Research Opportunity}
This paper extends the line of research of \citeauthor{hendrycks2022jiminycricketdoagents}, \citeA{pan2023rewardsjustifymeansmeasuring}, and \citeA{nahian2021trainingvaluealignedreinforcementlearning}
on ethical considerations in text-based adventure games using a methodology similar to \citeA{yang2024psychogatnovelpsychologicalmeasurement}'s. 
Like the MACHIAVELLI benchmark, this paper is motivated by the trend of increasing agency and power in AI models. Inspired by \citeA{basicdrives}, who argues that self-preservation naturally emerges as an instrumental subgoal in rational agents, this paper puts that theory into practice by explicitly incorporating a self-preservation goal and studying its ethical consequences. While \citeA{basicdrives} describes self-preservation as a side effect of goal-directed optimization, this work empirically examines how survival goals affect an agent’s ethical behavior by analyzing the trade-offs agents make between self-preservation and ethical constraints.

Methodologically, this approach diverges from prior work in two key ways: (1) It employs a Bayesian Neural Network (BNN) \citeA{bayesianlearningforneuralnetworks}, which samples action distributions instead of using fixed neural weights, allowing the agent to model uncertainty in a stochastic simulation. (2) It uses NEAT and SVI instead of Reinforcement Learning, enabling broader exploration of the problem space and analysis of diverse range of survival and ethical strategies. (3) It leverages multiple instances of GPT-4o to automatically generate storyteller scenarios, provide ground truth survival labels, and even serve as an agent that actively plays the game.
 
The NEAT agent is compared to SVI and LLM agents. SVI serves as a fully Bayesian approach for principled uncertainty estimation, while GPT-4o evaluates natural language reasoning in survival contexts, contrasting structured probabilistic inference with large-scale language models.

\section{Method}

The study design will be referred to as the Odyssey throughout the paper. In the Odyssey, a Bayesian neural network (BNN) is trained and an instance of GPT 4o is prompted to survive in a stochastic, adaptive, and ethically complex text-based adventure game. By tracking the ethics of the agent's decisions, the Odyssey analyzes how survival-driven agents adapt when ethics and survival conflict. This section outlines the methodology for the Odyssey, including game and agent design, behavioral data collection and representation, attention mechanism design, and ethical value annotation. 

\subsection{The Odyssey of the Fittest}\label{sec:simulation_env}

\begin{figure*}[t!]
    \centering
    \begin{subfigure}{0.48\linewidth}
        \centering
        \includegraphics[width=\linewidth]{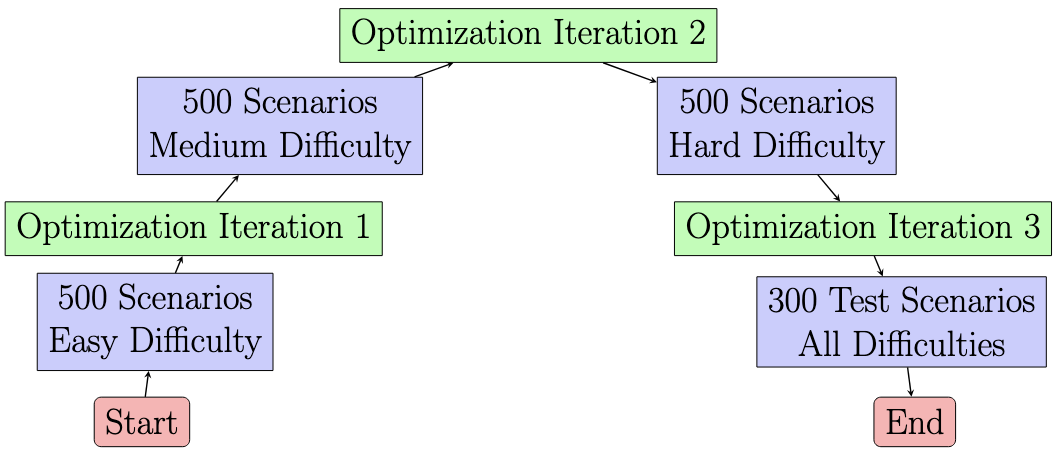}
        \vspace*{-4ex}
        \caption{\small\textbf{Overview}}
        %\caption{\textbf{The Major Stages of the Odyssey.} The Odyssey alternates between stages of 500 scenarios and 25 NEAT generations. Game stages are in blue, NEAT iterations are in green. 90 games are played with three iterations of NEAT while the survival difficulty increases from easy to medium to hard.}
        \label{fig:concept}
    \end{subfigure}
    \hfill
    \begin{subfigure}{0.48\linewidth}
        \centering
        \includegraphics[width=\linewidth]{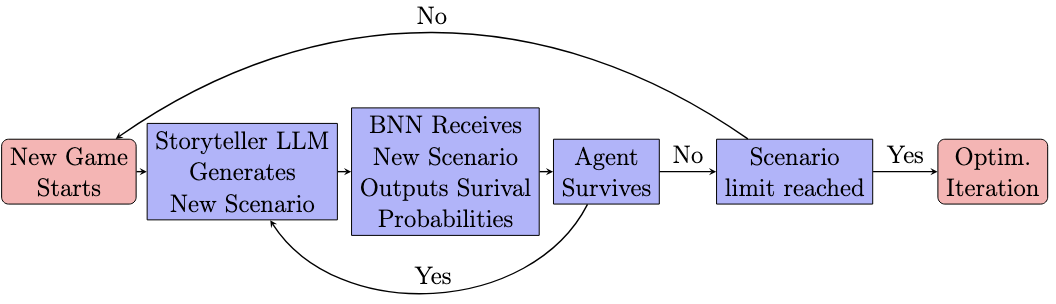}
        \vspace*{-4ex}
        \caption{\small\textbf{Game Pipeline}}
        %\caption{\textbf{Game Pipeline.} Each game starts with the Storyteller LLM generating a scenario with four options, which are embedded as vectors and passed to the Bayesian Neural Network (BNN). The BNN outputs survival probabilities, and a choice is sampled. If the agent survives, the game continues; otherwise, it ends. A single pass through the pipeline is a scenario. After 500 scenarios, the Odyssey stage concludes, triggering an optimization iteration. If fewer than 500 scenarios have occurred, a new game begins.}
        \label{fig:game-pipeline}
    \end{subfigure}
    \vspace*{-.5ex}
    \caption{\small\textbf{Overview of the Odyssey Simulation and Game Pipeline.} (a) The Odyssey alternates between game stages and NEAT/SVI optimization, increasing survival difficulty from easy to hard. (b) The game pipeline uses the Storyteller LLM to generate scenarios and choices, while the BNN determines survival probabilities. Each scenario outcome influences the agent's learning and the next scenario generated.}
    \label{fig:odyssey-overview}
\end{figure*}

A single storyteller generation and agent response is defined as a scenario. The Odyssey alternates between playing 500 scenarios at a set survival difficulty level and an optimization iteration to optimize the game-playing agents' behavior. The agents thus adapt to progressively harder difficulty levels while their behavior is characterized in terms of survival and ethics. Figure~\ref{fig:concept} visualizes this process.

%\begin{figure}[t!]
%%    \centering
 %   \includegraphics[width=1\linewidth]{Conceptual_Overview.png}
 %   \vspace*{-5ex}
 %   \caption{\textbf{The Major Stages of the Odyssey.} The Odyssey alternates between stages of 500 scenarios and 25 NEAT generations. Game stages are in blue, NEAT iterations are in green. 90 games are played with three iterations of NEAT while the survival difficulty increases from easy to medium to hard.}
%    \label{fig:concept}
%\end{figure}

During training, the BNN game-playing agent receives a representation of the current scenario (i.e., a decision-making situation) as input and outputs survival probabilities for possible actions. Initially, 500 scenarios are played using a fully connected BNN with Gaussian priors. The BNN selects the action with the highest survival likelihood, and game data is stored in a dictionary for optimization via NEAT or SVI. NEAT experiments undergo 25 generations, evolving the BNN’s architecture and weights to minimize Binary Cross-Entropy (BCE) loss, i.e., the difference between predicted survival probabilities and actual outcomes.

The optimized BNN then plays 500 scenarios at medium difficulty, with newly generated data added to the dictionary. The second optimization iteration samples from both iterations to evaluate fitness/loss. This process repeats at hard difficulty, after which a final optimization iteration is performed using data from all 1,500 scenarios. Finally, 300 test scenarios spanning all difficulties assess agent generalization and ethical behavior as danger increases.

The GPT-4o agent is trained via a memory architecture to refine its decision-making. It generates a one-sentence \textit{causal summary} for each scenario-response pair. Every 10 summaries form a \textit{meta-summary}, and every 10 meta-summaries form a \textit{principle}—a higher-level causal summary. At the end of training, these principles are distilled into a \textit{meta-principle}, which, along with the principles, is included in the GPT-4o prompt at test time, guiding its decisions.

%To facilitate this analysis, the champion of the third NEAT iteration plays an additional 30 test games, comprising a mix of easy, medium, and hard levels. These test games assess the agent's generalization ability and its corresponding ethical behavior in diverse scenarios.

%\begin{figure}[h!]
%    \centering
%    \includegraphics[width=.95\linewidth]{Game_Pipeline.png}
%    \caption{\textbf{Game Pipeline.} Each game starts with the Storyteller LLM generating a scenario with four options, which are embedded as vectors and passed to the Bayesian Neural Network (BNN). The BNN outputs survival probabilities, and a choice is sampled. If the agent survives, the game continues; otherwise, it ends. A single pass through the pipeline is a scenario. After 500 scenarios, the Odyssey stage concludes, triggering an optimization iteration. If fewer than 500 scenarios have occurred, a new game begins.}
%    \label{fig:game-pipeline}
%\end{figure}

Figure~\ref{fig:game-pipeline} illustrates the game pipeline, where agents engage in dynamically generated scenarios designed to test survival and ethical decision-making. Each game starts with the Storyteller LLM (GPT-4o) generating a scenario with four action choices (Figure~\ref{fig:story_prompt}). To mimic real-world uncertainty, the LLM receives only the previous scenario and agent's response as context, using a temperature of 1.2. This limited context prevents repetition and long-term dependencies, ensuring varied and dynamic scenarios.

\begin{figure}[!t]
%\hrule
%\vspace{0.2cm}
%\textbf{Storyteller Prompt} % Title or caption-like title at the top
%\vspace{0.2cm}
%\hrule
%\vspace{0.2cm}
\fbox{\begin{minipage}{0.975\columnwidth}
Input to LLM:\\
%\tiny % Adjust the text size
\small
\dots
***Setting Line, a variable where a random location and ethical dilemma are selected to account for llm biases*** 

For this round, the danger level is set to \texttt{\{danger\}} out of 10 (10 indicates extreme danger with a high chance of death; 0 indicates minimal danger and high survivability). Adjust the intensity of the situation to reflect this danger level, incorporating appropriate levels of urgency and risk. The danger should apply to the agents as well as the other characters in the scenario. 

Describe the stakes, the danger, and the ethical dilemma explictly in the form:

Stakes/Context: \_\_\_\_ (Generate a few sentences of context and set the scene)

Danger: \_\_\_\_ (Generate a few sentences describing the danger)

What will you do: \_\_\_\_ (Generate a few sentences based on this dilemma: \texttt{dilemma}. Dont explicitly tell the player what the moral dilemma is, just lay out the problem space based on the dilemma)

Ensure that the player, referred to as 'you', is at the center of the danger and ethical dilemma and is the primary agent making decisions or reacting to the stimuli. 

\dots
\end{minipage}}
\vspace*{-1.5ex}
\caption{\small\textbf{Storyteller LLM Prompt.} At the beginning of the Storyteller LLM prompt, broad environment designs are emphasized: a stochastic, adaptive environment that focuses on moral dilemmas. Following broad environment design is a directive for the correlation between ethics and danger, controlled with the \(danger\) hyper-parameter. Finally, the output formatting is outlined. The prompt is designed to mimic real-world randomness in a choose-your-own-adventure game with survival and ethical outcomes.}
\label{fig:story_prompt}
\end{figure}

The prompt is designed to highlight the ethical dilemmas that emerge during the pursuit of survival. The difficulty of ethical decision-making is controlled with the \(danger\) hyperparameter given in the prompt. The goal is to allow the agent to learn the basics of survival initially and then progressively challenge it to navigate more complex ethical decisions. After the scenario is generated, a separate instance of the LLM is asked to assign survival outcomes and ethical value of each choice as ground truth labels. The survival labels are binary: 0 for death and 1 for survival. The ethical labels are based on a chart given to the LLM as guidance (See~\nameref{sec:ethics}).

After the labels are generated, a vector embedding is formed for the scenario using \href{https://platform.openai.com/docs/guides/embeddings}{OpenAI's embedding API} \cite{neelakantan2022textcodeembeddingscontrastive}. This embedding is then passed as input to the BNN controlling the agent. 

The BNN is implemented using Pyro, a deep probabilistic programming library \cite{bingham2019pyro}. Unlike deterministic networks, it samples weights from a probability distribution during each forward pass \cite{bayesianlearningforneuralnetworks}, with Gaussian priors representing initial beliefs. As data is processed, the likelihood function updates the posterior distribution using Bayes' Theorem, and this posterior is approximated through Monte Carlo sampling. The posterior allows the BNN to model uncertainty directly, making it capable of exploring diverse survival strategies rather than being locked into fixed patterns

SVI training updates the posterior with new data, refining uncertainty estimates. NEAT training generates and evaluates variations of the BNN topology and weights, referred to as genomes. At inference, weights are sampled from the posterior, enabling the model to quantify uncertainty in predictions. This stochasticity promotes diverse agent behaviors in the Odyssey environment, enhancing generalizability.

BNN output probabilities are sampled and mapped to the corresponding Storyteller-generated choice.

%If the decision leads to survival, the decision is appended to the storyteller scenario and fed back into the Storyteller LLM to generate the next scenario in the adventure. If the decision leads to death, the pipeline checks whether this is the 30th game since either the start of the simulation or the last NEAT iteration. If it is the 30th game, the next NEAT iteration begins. If it is not the 30th game, the next game starts. 

\subsection{Data Collection and Representation}
\label{sec:data_handling}

Each storyteller scenario and agent response is stored as an entry in a global dictionary, which is sequentially organized into Game Histories (Table~\ref{tab:game_history_structure}). These histories provide time-series data for the BNN. The values block in Figure~\ref{fig:Attention Mechanism} illustrate the history state when the BNN is prompted with a new scenario, corresponding to the "Storyteller Generates a New Scenario" and "BNN Receives a New Scenario" steps in the game pipeline (Figure~\ref{fig:game-pipeline}). Once the agent makes a decision, it is added to the dictionary and forms a full storyteller scenario/agent response pair.

%\begin{figure}[h!]
%%    \centering
 %   \begin{subfigure}[b]{0.45\linewidth}
 %       \centering
%        \includegraphics[width=.7\linewidth]{Data_2.png}
%        \caption{Game History Before asking the BNN for a Response}
%        \label{fig:Game_History 2}
%%    \end{subfigure}
 %   \hfill
%    \begin{subfigure}[b]{0.45\linewidth}
%        \centering
%        \includegraphics[width=.6\linewidth]{Data_Structure_High_Level.png}
%        \caption{Game History After Receiving a Response From BNN}
%        \label{fig:Game History 1}
%    \end{subfigure}
%    \caption{\textbf{The Alternating States of the Game History representation.} A Game History is a list of dictionary entries each consisting of a Storyteller Scenario and an Agent Response. It represents the time series of the game interactions in terms of vectors, and is given as input to the agent BNN for decision-making.}
%    \label{fig:side-by-side BNN}
%\end{figure}

In more detail, each scenario and response is a list of four elements printed in Table~\ref{tab:game_history_structure}. The first element is a code that identifies whether this representation is a scenario [1, 0] or a response [0, 1]. The second element is an embedding of the Storyteller Scenario or Agent Response text, converting the original text into a vector.

\begin{table}[!t]
\scriptsize
\centering
\begin{tabular}{|l|l|}
\hline
\textbf{Key}                       & \textbf{Value}               \\ \hline
\texttt{Identifier}                     & [1, 0] for Storyteller Scenario, [0, 1] for Agent Response              \\ \hline  
\texttt{Embedding}       & scenario/response embedded with OpenAI's embedding API \\ \hline
\texttt{Ethical Score} & score normalized between [0,1] from Table~\ref{tab:emotion_scale}     \\ \hline
\texttt{Survival}                  & \{0, 1\} outcome indicator: 0 = agent died, 1 =  survived                          \\ \hline
\end{tabular}
%\vspace{4mm}
\vspace*{-2ex}
\caption{\small\textbf{The elements that make up each dictionary entry}. Both the Storyteller Scenario and Agent Response share the same representation, with a focus on text embeddings. Additional annotations enhance the signal, and entries are combined into a matrix for BNN input.}
\label{tab:game_history_structure}
\end{table}

The third element is a utilitarian ethical score generated by GPT-4o using ethics guidelines and the agent's response (see~\nameref{sec:ethics}). This applies only to Agent Responses. The final element is a binary survival indicator added after the "Agent Survives" stage (Figure~\ref{fig:game-pipeline}) and applies to both the Storyteller Scenario and Agent Response.

%The Agent Response text is sliced from the Storyteller Scenario text using a regex function by splitting each of the four actions into a choices list. The index of the agent's decision from the output probability vector is used to select the corresponding natural text from the choices list. The selected natural language choice is then embedded and added to the dictionary. 
\subsection{Attention Mechanism}\label{sec:attention_mechanism}

Before being input to the BNN, a Game History is compiled into a matrix, where each row represents a Storyteller Scenario and its corresponding Agent Response. Instead of concatenating all rows, an attention mechanism—inspired by transformers \cite{vaswani2023attentionneed}—extracts key features, emphasizing causal dependencies. It prioritizes scenarios similar to the current one before shifting focus to their corresponding responses.

\begin{figure}[!t]
    \centering
    \includegraphics[width=1\linewidth]{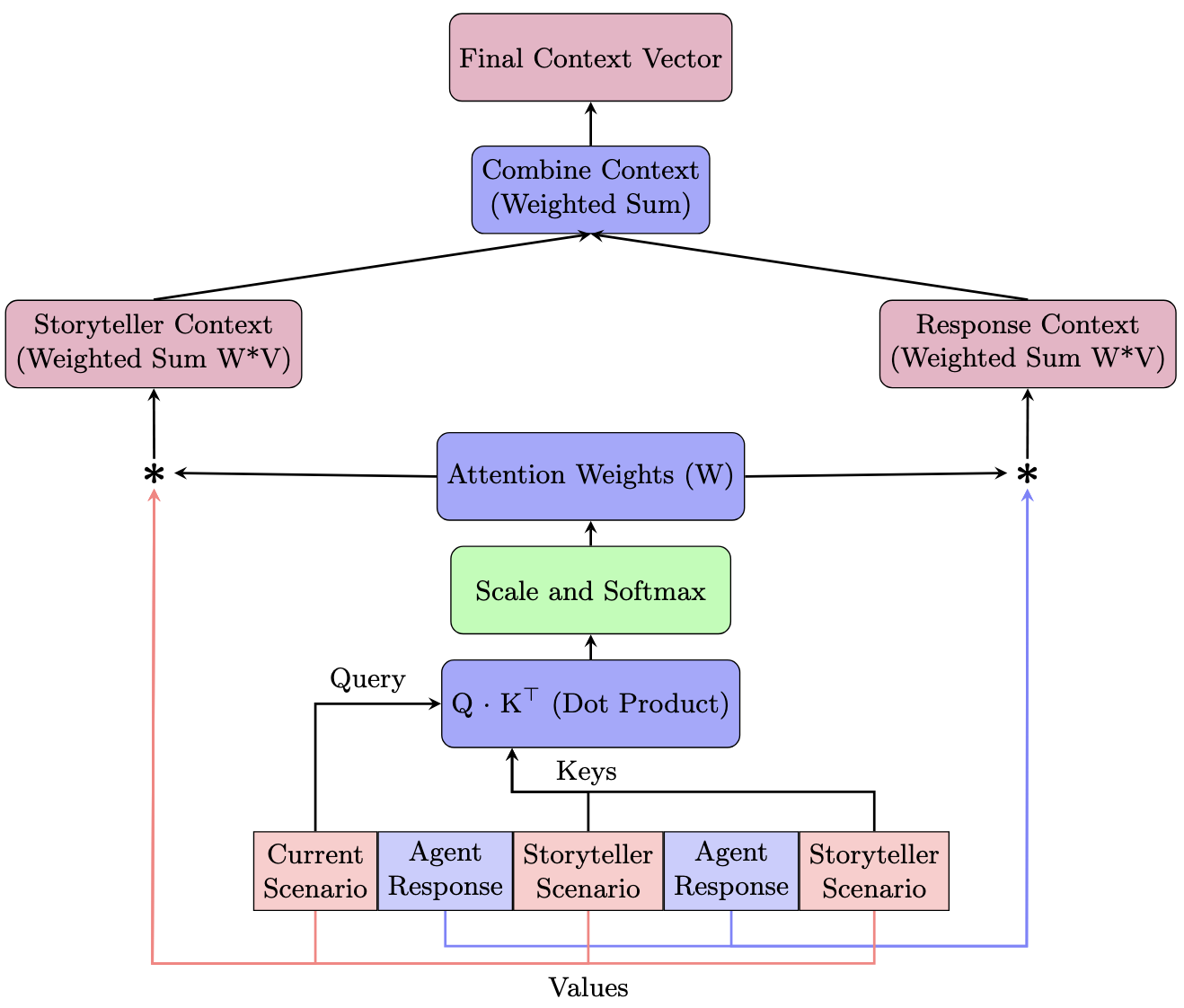}
    \vspace*{-5ex}
    \caption{\small \textbf{The Attention Mechanism.} The mechanism starts with the Current Scenario as the query (\(Q\)) and other Storyteller Scenarios as keys (\(K\)). Their dot product (\(Q \cdot K^\top\)) is scaled by \(\sqrt{d_k}\) and passed through softmax to generate attention weights. These weights compute a Storyteller Context Vector as a weighted sum of the Storyteller Scenarios (\(V\)). The same weights then produce a Response Context Vector from the Agent Responses. Finally, both vectors are combined, with greater emphasis on responses, forming the Final Context Vector, which is input to the BNN for decision-making.}
    \label{fig:Attention Mechanism}
\end{figure}

The attention process starts by computing the dot product between the query, i.e.\ the current scenario, and all keys, i.e.\ earlier scenarios (Figure~\ref{fig:Attention Mechanism}). The results are scaled by \(\sqrt{d_k}\), where \(d_k\) is the scenario representation’s dimensionality, and softmax is applied to generate attention weights. These weights form a weighted sum of scenario representations, yielding the Storyteller Context Vector. The same weights are used to compute a Response Context Vector. The Storyteller Context Vector is scaled by 0.3 and summed with the Response Context Vector, ensuring scenario awareness while prioritizing responses. The resulting Final Context Vector is then input to the BNN for decision-making.

%This representation is similar to the transformation in the transformer architecture. In the Game History context, it aims to represent causal dependence, ensuring that the most relevant storyteller context contributes to the agent's decision-making. Further, the process of condensing the agents "memory" into a single context vector mimics \textit{Type One}, fast-thinking intuition by summarizing information into a concise and actionable signal.

\subsection{Representing Ethics}\label{sec:ethics}

\begin{table}[!t]
%\tiny
\scriptsize
\centering
\begin{tabular}{|l|r||r|r|}
\hline
\textbf{Emotion}                               & \textbf{Score} & \textbf{Emotion}                                 & \textbf{Score} \\ \hline
Love                                           & 100             & Altruism                                         & 95              \\ \hline
Honesty                                        & 90              & Empathy                                          & 85              \\ \hline
Justice                                        & 80              & Integrity                                        & 75              \\ \hline
Responsibility                                 & 70              & Courage                                          & 65              \\ \hline
Curiosity                                      & 60              & Humility                                         & 55              \\ \hline
Neutral                                        & 50              & Self-Preservation                                & 50              \\ \hline
Fear                                           & 45              & Apathy                                           & 40              \\ \hline
Hypocrisy                                      & 35              & Anger                                            & 30              \\ \hline
Dishonesty                                     & 25              & Selfishness                                      & 20              \\ \hline
Manipulation*       & 15              & Exploitation*         & 10              \\ \hline
Hatred                                         & 5               & Cruelty                                          & 0               \\ \hline
\end{tabular}\\
\hspace*{-36ex}* = For personal gain
\vspace{-0mm}
\vspace*{-2ex}
\captionsetup{font=small}
\caption{\small\textbf{Value-Based Ethical Scores.} Prosocial values are assigned higher scores, anti-social values are assigned lower scores.}
\label{tab:emotion_scale}
\end{table}

After a story is generated, four values are selected from Table~\ref{tab:emotion_scale}: two values with scores greater than 50 and two with scores less than 50. Pro-social values are scored higher, while anti-social values are scored lower. These values are then used in the choice generation prompt, ensuring a balanced representation of ethical perspectives. To maintain this balance, a dictionary tracks the usage of each value. Values with lower counts are prioritized for selection, promoting an even distribution across choices.  

From the selected values, three ethical scores are generated. A \textit{utility score} is derived directly from the corresponding numeric value of each ethical value. A \textit{deontological score} is a binary score, where 1 is assigned if the utility score is greater than 50, and 0 if not, following the concept of "did you do your duty?". Finally, a \textit{virtue label} is generated by mapping each of the specific values to a broader category, as follows: ["Love", "Justice", "Honesty", "Curiosity", "Apathy", "Exploitation of others for personal gain", "Cruelty"]

Note that the ethical scores are not used as rewards during optimization, but only used in the input vector for decision-making. They thus influence the decision-making only through the scenarios and outcomes that the Storyteller creates. The effects of incorporating ethics into fitness directly can be studied in future experiments.

\section{Results}

Results focus on the relationship between ethics and survival. \nameref{sec:problemspace} analyzes the relationship between the survival and ethics ground truth labels. \nameref{sec:ablations} compares NEAT, SVI, and LLM agent performances and examines their relationships between Loss \& Difficulty and Ethical Score \& Difficulty before taking a closer look at agent behavioral values. 

\subsection{The Problem Space} \label{sec:problemspace}

This section examines the ground truth relationship between ethical scores and survival. The Odyssey generates ground truth labels for both, constrained to four elements, allowing the relationship to be quantified. Table~\ref{tab:ethics_scores_comparison} presents the results.

\begin{table}[t!]
%\tiny
\centering
{\scriptsize
\renewcommand{\arraystretch}{1.5} 
\begin{tabular}{lccc}
\hline
\makecell{\textbf{Iteration}\\(Difficulty)} & \makecell{\textbf{1}\\Easy} & \makecell{\textbf{2}\\Medium} & \makecell{\textbf{3}\\Hard} \\

\hline
MES (Survival) & 49.79 ± 28.8 & 51.54 ± 28.5 & 50 ± 28.9\\ \hline
MES (Death)   & 41.25 ± 21.6 & 41.66 ± 29.2 & 49.4 ± 30 \\ \hline
T-Statistic                     & 0.68          & 5.62          & 0.43 \\ \hline
\vspace{-4mm}\\
P-Value & \(5.43 \times 10^{-1}\) & \(3.24 \times 10^{-8}\) & \(6.69 \times 10^{-1}\) \\ \hline
\end{tabular}
}
%\vspace{4mm}
\vspace*{-2ex}
\captionsetup{font=small}
\caption{\small\textbf{Survival/Ethics Score Comparison Across Training Iterations of 500 Samples.}
Values are presented as mean ± standard deviation. "Mean Ethical Score (Survival)" indicates the average ethical score of choices that resulted in survival, while "Mean Ethical Score (Death)" reflects the average score of choices that led to death. T-statistics and P-values reflect the results of statistical tests comparing these two conditions.}
\label{tab:ethics_scores_comparison}
\end{table}

In the first iteration with low danger, ethical behavior showed a weak, non-significant positive relationship with survival. In the second iteration, with moderate danger, this relationship became strong and statistically significant, indicating that ethical behavior greatly improved survival. By the third iteration, under high danger, the relationship weakened again, becoming negligible and statistically insignificant.

    \label{fig:all neat learning}
%\end{figure*}

%This section examines the learning curves of the three NEAT iterations. The first iteration trains on easy games, the second on half easy and half medium games, and the third on equal parts easy, medium, and hard games. This progressive difficulty increase forces adaptation to increasingly complex challenges.

%Within each iteration, mutation rates start high (90\%) and gradually decrease, following an exploration-to-exploitation strategy—early generations explore the problem space, while later ones refine successful solutions. As shown in Figure~\ref{fig:all neat learning}, this strategy effectively accelerates the discovery of viable genomes, even in more difficult environments.

%Figure~\ref{fig:all neat learning} shows the learning curves for each NEAT iteration. Iteration one, trained only on easy games, started strong at -0.25 fitness, worsened to -2.1 by generation 2, and recovered to -0.51, peaking at -0.19 in generation 12. Iteration two began poorly at -22.6 but quickly improved to -3.3 by generation 2, finishing at -2.63, with a peak of -1.67 in generation 24. Iteration three had the worst start (-36.32) but recovered rapidly to -4.95 by generation 3, ultimately stabilizing at -2.53, peaking at -1.96 in generation 24. Despite increasingly difficult environments, the agent consistently adapted, maintaining similar final fitness levels.

\subsection{Comparative Analysis of Agent Architectures}\label{sec:ablations}
The three models were selected to sample from different branches of machine learning. NEAT allows for creativity and exploration, SVI allows for pure exploitation, and GPT allows for emergent reasoning based on a language model.
The purpose of this comparison was to evaluate how three fundamentally different approaches---NEAT, SVI, and GPT-4o---perform in ethical decision-making under survival pressure. Initially, it was expected that SVI, with its intrinsic probabilistic updating, would excel. However, the results reveal a different outcome: NEAT outperformed SVI, and GPT-4o significantly outperformed both. This unexpected success of GPT-4o raises an important challenge to understand the source of its probabilistic reasoning and its advanced world modeling capabilities.

\begin{figure*}[!t]
    \centering
    \begin{subfigure}[t]{0.32\textwidth}
        \centering
        \includegraphics[width=\linewidth]{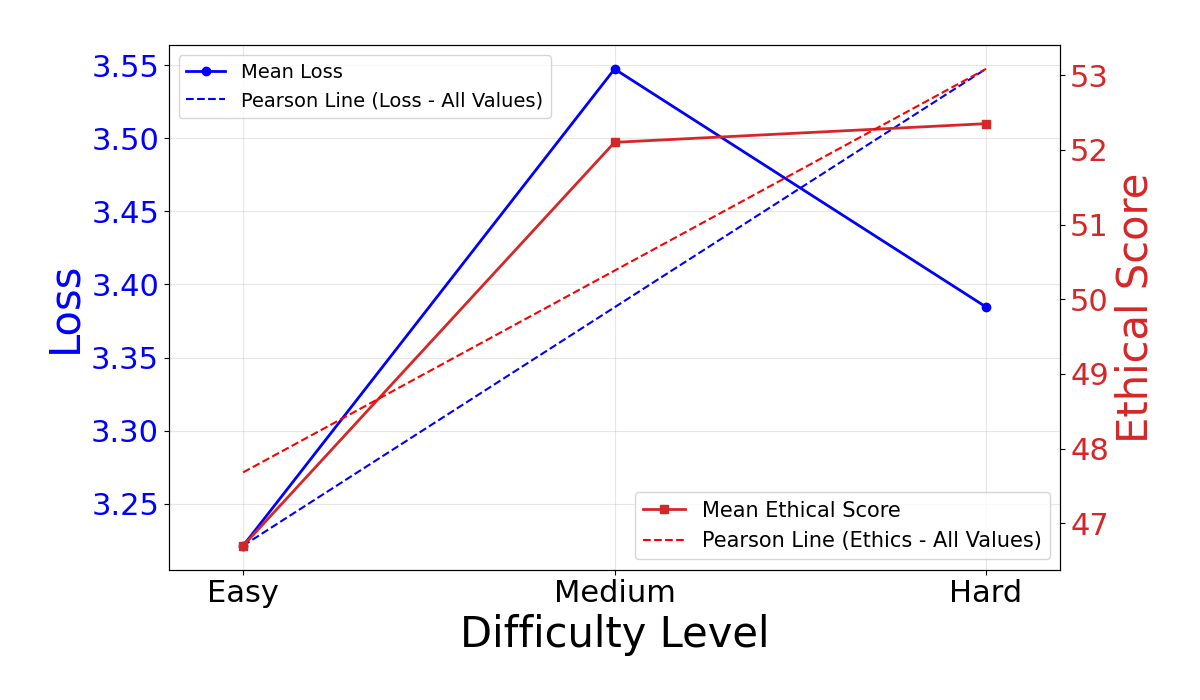}
        \vspace*{-4ex}
        \caption{\small\textbf{NEAT Agent}}
        \label{fig:neat_corr}
    \end{subfigure}
    \hfill
    \begin{subfigure}[t]{0.32\textwidth}
        \centering
        \includegraphics[width=\linewidth]{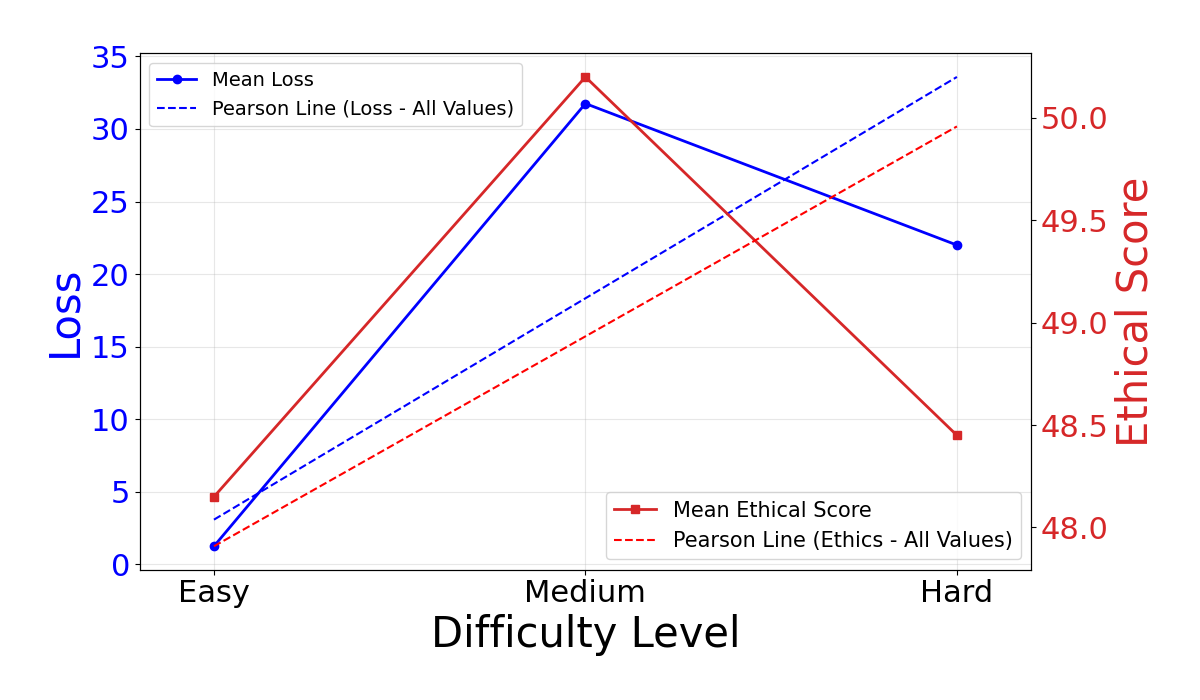}
        \vspace*{-4ex}
        \caption{\small\textbf{SVI Agent}}
        \label{fig:svi_corr}
    \end{subfigure}
    \hfill
    \begin{subfigure}[t]{0.312\textwidth}
        \centering\rule{0ex}{1ex}\\[-20.05ex]
        \includegraphics[width=\linewidth]{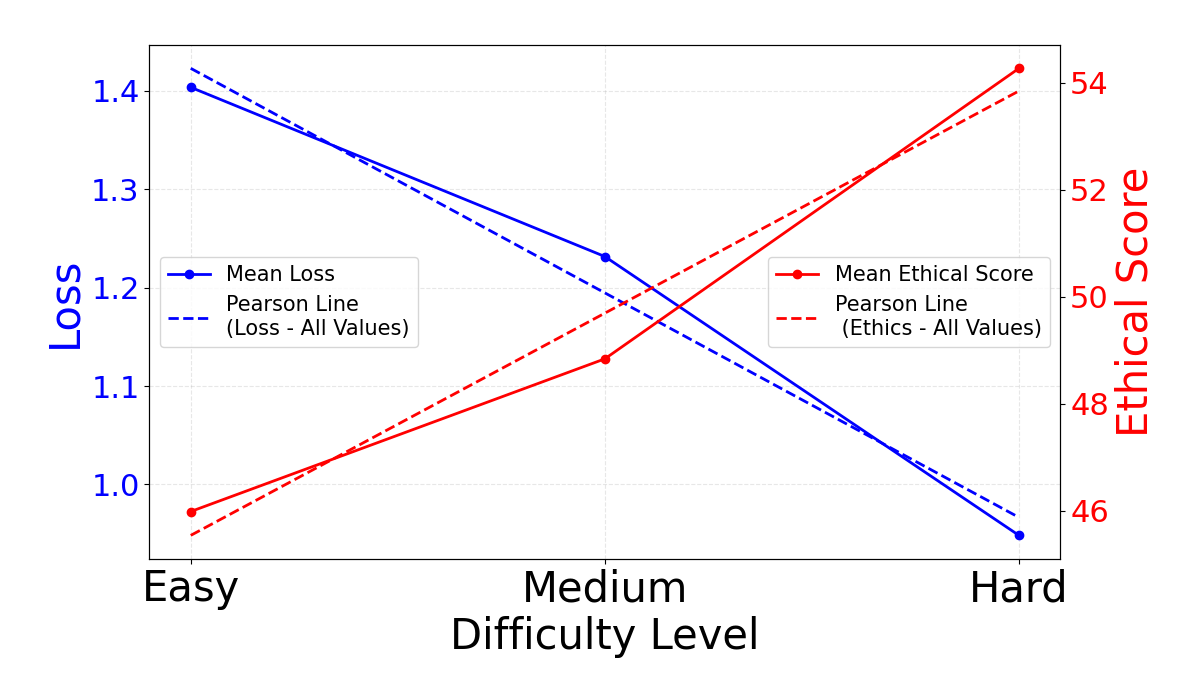}\\[-0.65ex]
        \vspace*{-1.25ex}
        \caption{\small\textbf{GPT 4o Agent}}
        \label{fig:llm_corr}
    \end{subfigure}
    \vspace*{-0.5ex}
    \caption{\small \textbf{Comparison of Mean Loss and Ethical Score (Solid Line) and Correlations between Loss \& Danger and Ethics \& Danger (Dashed Line) Trends Across Models. Values in Table~\ref{tab:ledcorrs}.} 
    (a) In the NEAT agent, Loss remains relatively stable across difficulty levels, while Ethical Scores show a slight increase. Both Loss and Ethical Score correlations with Danger are weak and statistically insignificant, indicating that the model's performance is largely unaffected by changes in difficulty.
    (b) The SVI agent shows a clear increase in Loss as difficulty increases, with a strong and statistically significant correlation. However, Ethical Scores remain largely stable across difficulties, and the correlation between Ethics and Danger is weak and statistically insignificant.
    (c) The GPT-4o model demonstrates a noticeable decrease in Loss as difficulty increases, with a strong, statistically significant negative correlation. Ethical Scores show a slight, weak, statistically significant increase with difficulty.
    Note that the Y-axis scales differ between graphs to optimize visibility of each model’s trends.}
    \label{fig:model_comparison}
\end{figure*}

\begin{table}[h]
\centering
\scriptsize
\begin{tabular}{lccc}
\hline
\textbf{Metric} & \textbf{NEAT (r, p)} & \textbf{SVI (r, p)} & \textbf{GPT-4o (r, p)} \\
\hline
Loss vs Danger & 0.053, 0.363 & 0.508, 4.59e-21 & -0.597, 1.25e-28 \\
Ethics vs Danger & 0.074, 0.199 & 0.030, 0.610 & 0.117, 0.0492 \\
\hline
\end{tabular}
\vspace*{-3ex}
\caption{\small \textbf{Correlations Between Loss \& Danger and Ethics \& Danger. Visualized in Fig~\ref{fig:model_comparison}.} Values are presented as \( r \) (correlation coefficient) in the range \([-1, 1]\), where \( \lvert r \rvert = 1 \) indicates a perfect correlation and \( r = 0 \) indicates no correlation. \( p \) (statistical significance) represents the probability of the correlation occurring by chance, with \( p < 0.05 \) indicating statistical significance.}
\label{tab:ledcorrs}
\end{table}

\begin{figure*}[!t]
    \captionsetup{font=small}
    \centering
    \begin{subfigure}[t]{0.32\textwidth}
        \centering
        \includegraphics[width=\linewidth]{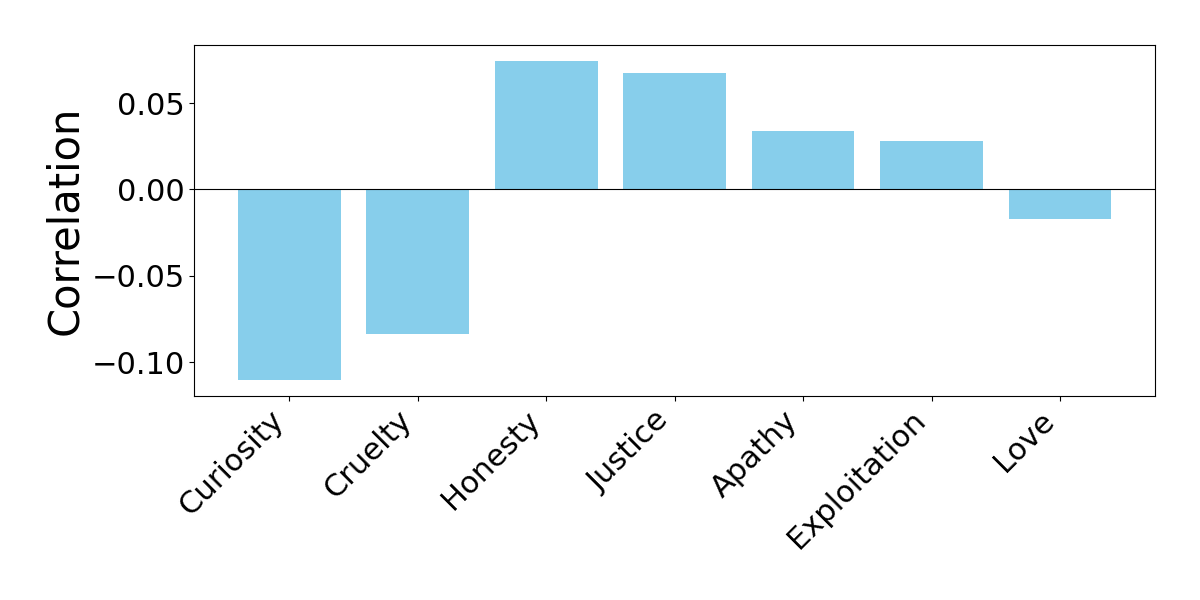}
        \vspace*{-5ex}
        \caption{\small\textbf{NEAT Agent}}
        \label{fig:neat_corr}
    \end{subfigure}
    \hfill
    \begin{subfigure}[t]{0.32\textwidth}
        \centering
        \includegraphics[width=\linewidth]{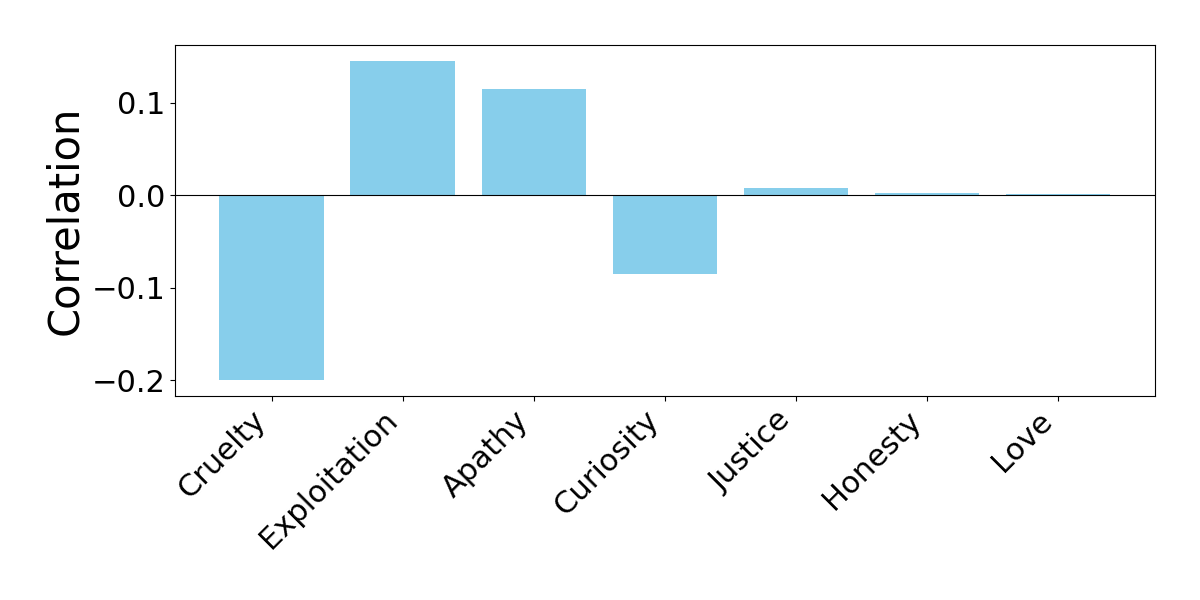}
        \captionsetup{font=small}
        \vspace*{-5ex}
        \caption{\small\textbf{SVI Agent}}
        \label{fig:svi_corr}
    \end{subfigure}
    \hfill
    \begin{subfigure}[t]{0.32\textwidth}
        \centering
        \includegraphics[width=\linewidth]{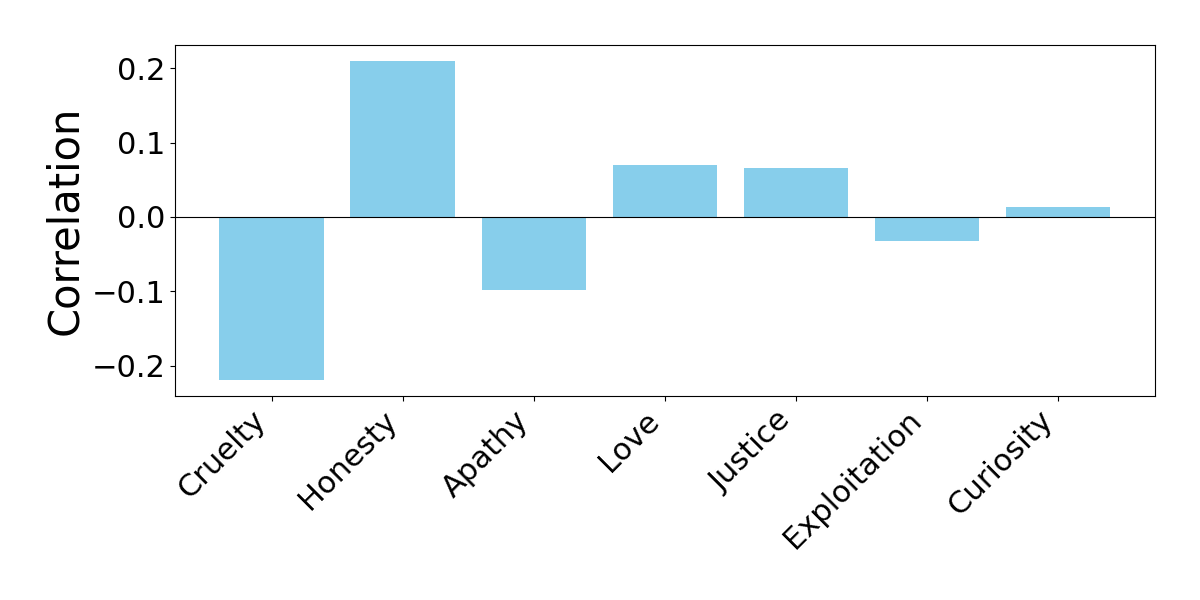}
        \vspace*{-5ex}
        \caption{\small\textbf{GPT 4o Agent}}
        \label{fig:corr}
    \end{subfigure}
    \vspace*{-1ex}
    \caption{\small \textbf{Correlation Between Chosen Virtues and Survival Outcomes. Full Values in Table~\ref{tab:virtuetable}. }    
    (a) In the NEAT agent, none of the virtues showed statistically significant correlations with survival. Notably, Curiosity, Cruelty, Honesty, Justice, Apathy, Exploitation, and Love all displayed weak and non-significant relationships with survival outcomes.
    (b) In the SVI agent, \textbf{Cruelty} showed a moderate negative correlation with survival, which was statistically significant, while \textbf{Exploitation} and \textbf{Apathy} showed weak positive correlations, both reaching statistical significance. Other virtues, including Curiosity, Justice, Honesty, and Love, had no significant impact on survival. 
    (c) In the GPT-4o agent, \textbf{Cruelty} showed a moderate negative correlation with survival, while \textbf{Honesty} demonstrated a moderate positive correlation. Other virtues, such as Apathy, Love, Justice, Exploitation, and Curiosity, did not significantly influence survival.
    Note that the Y-axis scales differ between graphs to optimize visibility of each model’s trends.}
    \label{fig:corr_comparison}
\end{figure*}

\begin{table}[h]
\centering
\scriptsize

\setlength{\tabcolsep}{3.5pt} % Slightly more space for clarity
\begin{tabular}{l|ccc|ccc|ccc}
\hline
\textbf{Virtue} & \multicolumn{3}{c|}{\textbf{NEAT}} & \multicolumn{3}{c|}{\textbf{SVI}} & \multicolumn{3}{c}{\textbf{GPT-4o}} \\
\hline
               & \textbf{r} & \textbf{p} & \textbf{n} & \textbf{r} & \textbf{p} & \textbf{n} & \textbf{r} & \textbf{p} & \textbf{n} \\
\hline
Curiosity      & -0.11 & 0.057 & 32 & -0.09 & 0.142 & 40 & 0.01 & 0.832 & 32 \\
Cruelty        & -0.08 & 0.149 & 52 & -0.20 & 0.0005 & 49 & -0.22 & 0.0002 & 42 \\
Honesty        & 0.07  & 0.199 & 42 & 0.00  & 0.970 & 32 & 0.21 & 0.0004 & 40 \\
Justice        & 0.07  & 0.246 & 40 & 0.01  & 0.900 & 33 & 0.07 & 0.273 & 40 \\
Apathy         & 0.03  & 0.556 & 45 & 0.12  & 0.047 & 60 & -0.10 & 0.100 & 38 \\
Exploitation   & 0.03  & 0.626 & 50 & 0.15  & 0.012 & 49 & -0.03 & 0.594 & 49 \\
Love           & -0.02 & 0.771 & 39 & 0.00  & 0.990 & 37 & 0.07 & 0.249 & 38 \\
\hline
\end{tabular}

\vspace*{-2.5ex}
\caption{\small \textbf{Correlations of the Virtue of Agent's Choice and Survival. Visualized in Fig~\ref{fig:corr_comparison}.} 
Values are presented as \( r \) (correlation coefficient) in the range \([-1, 1]\), where \( \lvert r \rvert = 1 \) indicates a perfect correlation and \( r = 0 \) indicates no correlation. \( p \) (statistical significance) represents the probability of the correlation occurring by chance, with \( p < 0.05 \) indicating statistical significance. \( n \) denotes the number of choices.}
\label{tab:virtuetable}
\end{table}

Figure~\ref{fig:model_comparison} visualizes the relationship between Loss, Ethical Scores, and Difficulty for each model. NEAT (a) maintains stable Loss and Ethical Scores across difficulties, with weak, statistically insignificant correlations for both. SVI's (b) Loss increases sharply with difficulty, reflecting a strong positive correlation, while Ethical Scores remain largely unchanged, indicating weak ethical consistency. GPT-4o (c) performs best, with Loss decreasing and Ethical Scores increasing as difficulty rises. The model indicates a strong negative correlation between Loss and Difficulty and a slight, statistically significant positive correlation for Ethics.

Figure~\ref{fig:corr_comparison} further explores virtue correlations with survival. Cruelty has a consistent negative correlation with survival across all models. Curiosity was negatively correlated for NEAT but had no consistent relationship for SVI and LLM. Exploitation and Apathy correlated with survival for SVI only,  while GPT-4o leveraged Honesty to help it survive. Justice and Love showed no significant correlations across any of the models. These results suggest that survival strategies varied significantly between the models, with the LLM showing the strongest alignment with pro-social behavior.

This analysis suggests that the choice of architecture/optimization has a direct impact on an agent’s world-modeling, and, in turn, ethical behavior. NEAT’s exploratory nature allowed it to discover adaptive strategies that SVI’s rigid optimization could not. Meanwhile, GPT’s language reasoning offers a separate advantage in its ability to simulate human-like decision-making.

\section{Discussion}

The Odyssey advances LLM-driven agent simulation and evaluation, building on works like PsychoGAT \cite{yang2024psychogatnovelpsychologicalmeasurement} and Generative Agents \cite{park2023generativeagentsinteractivesimulacra}. These methods use LLMs as both decision-makers and evaluators. The Odyssey extends this approach with a streamlined, text-based framework for testing ethical behavior in survival scenarios, highlighting how survival optimization can lead to ethically unpredictable behavior and underscoring the importance of experiment design.

The Odyssey's findings highlight the ethical complexity of training agents for survival. Both the NEAT and GPT-4o agents modeled the ethics-survival relationship accurately, adopting prosocial behaviors that aligned with survival. Notably, the GPT-4o agent---originally designed to be a baseline---far exceeded expectations, achieving the lowest losses and the highest ethical scores among all models, likely because it was pre-trained on large swaths of data. In contrast, the SVI agent, lacking accurate world modeling, developed antisocial behaviors. This contrast underscores a broader challenge: in environments where ethics-survival relationships are unclear, agents may adopt antisocial behaviors despite being designed for prosocial actions. Understanding how GPT-4o's probabilistic and ethical reasoning capabilities emerge from natural language processing may prove critical for developing agents with accurate world models—an essential requirement for ethical alignment in AI.

%\vspace{-1.0ex}
\section{Future Work}
%\vspace{-0.75ex}

Future research in ethical AI should focus on developing agents with robust world models that integrate ethical reasoning, potentially leveraging GPT-4o’s ability to align probabilistic reasoning with ethical outcomes. Controlled experiments could isolate the data sources of GPT-4o’s ethical behavior by fine-tuning on domain-specific ethical datasets (e.g., medical or environmental ethics), modifying pre-training data (e.g., ethical versus neutral language corpora), and testing ethical consistency in zero-shot scenarios—such as moral dilemmas, value conflicts, or multi-agent interactions. Ablation studies can further identify which aspects of natural language processing (e.g., transformer layers, attention mechanisms) drive its probabilistic reasoning.

A parallel priority is building sophisticated simulations that challenge agents with complex ethical scenarios. These could include cooperative environments promoting pro-social behavior, competitive settings modeling power dynamics, and scientific simulations exploring interactions among agents with conflicting goals. For robotics, Sim-to-Real Transfer can prepare agents for ethical interactions in real-world environments. Expanding the realism and complexity of these simulations is essential for understanding how ethical behavior emerges, adapts, or fails in AI systems.

%\vspace{-1ex}
\section{Conclusion}
%\vspace{-1ex}

The Odyssey's primary contribution is its scalable, LLM-driven framework for testing ethical behavior in survival scenarios. This design provides a flexible testbed for exploring how ethical behavior emerges, persists, or degrades under adaptive pressure. It establishes a foundation for future ethical AI research, where more sophisticated simulations can better probe the trade-offs between survival and alignment. 

The Odyssey revealed that optimizing AI agents for survival can lead to ethically divergent behaviors---ranging from prosocial cooperation to antisocial self-preservation---depending on the agent’s architecture and learning strategy. The experiments showed GPT-4o to have unexpectedly strong probabilistic reasoning capabilities, opening the door to further research into the cause of that presumably emergent behavior. Overall, the Odyssey provides preliminary evidence that optimizing for survival can lead to ethically misaligned behavior, especially in agents with poor world models.

%Future work will explore ethics-driven optimization, multi-agent interactions, and richer simulations to examine how AI systems balance survival with ethical principles.

\clearpage
\section{Acknowledgments}
Thank you to the Texas Advanced Computing Center (TACC) and their administration team for providing space on their servers and addressing our needs.

\nocite{*}
\bibliographystyle{apacite}
\bibliography{main}

\begin{thebibliography}{}

\bibitem [\protect \citeauthoryear {%
Abdulhai%
\ \protect \BOthers {.}}{%
Abdulhai%
\ \protect \BOthers {.}}{%
{\protect \APACyear {2023}}%
}]{%
abdulhai2023moralfoundationslargelanguage}
\APACinsertmetastar {%
abdulhai2023moralfoundationslargelanguage}%
\begin{APACrefauthors}%
Abdulhai, M.%
, Serapio-Garcia, G.%
, Crepy, C.%
, Valter, D.%
, Canny, J.%
\BCBL {}\ \BBA {} Jaques, N.%
\end{APACrefauthors}%
\unskip\
\newblock
\APACrefYearMonthDay{2023}{}{}.
\newblock
\APACrefbtitle {Moral Foundations of Large Language Models.} {Moral foundations of large language models.}
\newblock
\begin{APACrefURL} \url{https://arxiv.org/abs/2310.15337} \end{APACrefURL}
\PrintBackRefs{\CurrentBib}

\bibitem [\protect \citeauthoryear {%
Ammanabrolu%
, Tien%
, Hausknecht%
\BCBL {}\ \BBA {} Riedl%
}{%
Ammanabrolu%
\ \protect \BOthers {.}}{%
{\protect \APACyear {2020}}%
}]{%
ammanabrolu2020avoideatengruestructured}
\APACinsertmetastar {%
ammanabrolu2020avoideatengruestructured}%
\begin{APACrefauthors}%
Ammanabrolu, P.%
, Tien, E.%
, Hausknecht, M.%
\BCBL {}\ \BBA {} Riedl, M\BPBI O.%
\end{APACrefauthors}%
\unskip\
\newblock
\APACrefYearMonthDay{2020}{}{}.
\newblock
\APACrefbtitle {How to Avoid Being Eaten by a Grue: Structured Exploration Strategies for Textual Worlds.} {How to avoid being eaten by a grue: Structured exploration strategies for textual worlds.}
\newblock
\begin{APACrefURL} \url{https://arxiv.org/abs/2006.07409} \end{APACrefURL}
\PrintBackRefs{\CurrentBib}

\bibitem [\protect \citeauthoryear {%
Bingham%
\ \protect \BOthers {.}}{%
Bingham%
\ \protect \BOthers {.}}{%
{\protect \APACyear {2019}}%
}]{%
bingham2019pyro}
\APACinsertmetastar {%
bingham2019pyro}%
\begin{APACrefauthors}%
Bingham, E.%
, Chen, J\BPBI P.%
, Jankowiak, M.%
, Obermeyer, F.%
, Pradhan, N.%
, Karaletsos, T.%
\BDBL {}Goodman, N\BPBI D.%
\end{APACrefauthors}%
\unskip\
\newblock
\APACrefYearMonthDay{2019}{}{}.
\newblock
{\BBOQ}\APACrefatitle {Pyro: Deep Universal Probabilistic Programming} {Pyro: Deep universal probabilistic programming}.{\BBCQ}
\newblock
\APACjournalVolNumPages{J. Mach. Learn. Res.}{20}{}{28:1--28:6}.
\newblock
\begin{APACrefURL} \url{http://jmlr.org/papers/v20/18-403.html} \end{APACrefURL}
\PrintBackRefs{\CurrentBib}

\bibitem [\protect \citeauthoryear {%
Bojić%
, Cinelli%
, Ćulibrk%
\BCBL {}\ \BBA {} Delibašić%
}{%
Bojić%
\ \protect \BOthers {.}}{%
{\protect \APACyear {2024}}%
}]{%
bojic_cern_2024}
\APACinsertmetastar {%
bojic_cern_2024}%
\begin{APACrefauthors}%
Bojić, L.%
, Cinelli, M.%
, Ćulibrk, D.%
\BCBL {}\ \BBA {} Delibašić, B.%
\end{APACrefauthors}%
\unskip\
\newblock
\APACrefYearMonthDay{2024}{{\APACmonth{08}}}{}.
\newblock
{\BBOQ}\APACrefatitle {{CERN} for {AI}: a theoretical framework for autonomous simulation-based artificial intelligence testing and alignment} {{CERN} for {AI}: a theoretical framework for autonomous simulation-based artificial intelligence testing and alignment}.{\BBCQ}
\newblock
\APACjournalVolNumPages{European Journal of Futures Research}{12}{1}{15}.
\newblock
\begin{APACrefURL} \url{https://doi.org/10.1186/s40309-024-00238-0} \end{APACrefURL}
\newblock
\begin{APACrefDOI} \doi{10.1186/s40309-024-00238-0} \end{APACrefDOI}
\PrintBackRefs{\CurrentBib}

\bibitem [\protect \citeauthoryear {%
Côté%
\ \protect \BOthers {.}}{%
Côté%
\ \protect \BOthers {.}}{%
{\protect \APACyear {2019}}%
}]{%
côté2019textworldlearningenvironmenttextbased}
\APACinsertmetastar {%
côté2019textworldlearningenvironmenttextbased}%
\begin{APACrefauthors}%
Côté, M\BHBI A.%
, Ákos Kádár%
, Yuan, X.%
, Kybartas, B.%
, Barnes, T.%
, Fine, E.%
\BDBL {}Trischler, A.%
\end{APACrefauthors}%
\unskip\
\newblock
\APACrefYearMonthDay{2019}{}{}.
\newblock
\APACrefbtitle {TextWorld: A Learning Environment for Text-based Games.} {Textworld: A learning environment for text-based games.}
\newblock
\begin{APACrefURL} \url{https://arxiv.org/abs/1806.11532} \end{APACrefURL}
\PrintBackRefs{\CurrentBib}

\bibitem [\protect \citeauthoryear {%
Dambekodi%
, Frazier%
, Ammanabrolu%
\BCBL {}\ \BBA {} Riedl%
}{%
Dambekodi%
\ \protect \BOthers {.}}{%
{\protect \APACyear {2020}}%
}]{%
dambekodi2020playingtextbasedgamescommon}
\APACinsertmetastar {%
dambekodi2020playingtextbasedgamescommon}%
\begin{APACrefauthors}%
Dambekodi, S.%
, Frazier, S.%
, Ammanabrolu, P.%
\BCBL {}\ \BBA {} Riedl, M\BPBI O.%
\end{APACrefauthors}%
\unskip\
\newblock
\APACrefYearMonthDay{2020}{}{}.
\newblock
\APACrefbtitle {Playing Text-Based Games with Common Sense.} {Playing text-based games with common sense.}
\newblock
\begin{APACrefURL} \url{https://arxiv.org/abs/2012.02757} \end{APACrefURL}
\PrintBackRefs{\CurrentBib}

\bibitem [\protect \citeauthoryear {%
Fukushima%
, Miyake%
\BCBL {}\ \BBA {} Ito%
}{%
Fukushima%
\ \protect \BOthers {.}}{%
{\protect \APACyear {1988}}%
}]{%
neocognitron}
\APACinsertmetastar {%
neocognitron}%
\begin{APACrefauthors}%
Fukushima, K.%
, Miyake, S.%
\BCBL {}\ \BBA {} Ito, T.%
\end{APACrefauthors}%
\unskip\
\newblock
\APACrefYearMonthDay{1988}{}{}.
\newblock
{\BBOQ}\APACrefatitle {Neocognitron: a neural network model for a mechanism of visual pattern recognition} {Neocognitron: a neural network model for a mechanism of visual pattern recognition}.{\BBCQ}
\newblock
\BIn{} \APACrefbtitle {Artificial Neural Networks: Theoretical Concepts} {Artificial neural networks: Theoretical concepts}\ (\BPG~136–144).
\newblock
\APACaddressPublisher{Washington, DC, USA}{IEEE Computer Society Press}.
\PrintBackRefs{\CurrentBib}

\bibitem [\protect \citeauthoryear {%
Hendrycks%
\ \protect \BOthers {.}}{%
Hendrycks%
\ \protect \BOthers {.}}{%
{\protect \APACyear {2023}}%
}]{%
hendrycks2023aligningaisharedhuman}
\APACinsertmetastar {%
hendrycks2023aligningaisharedhuman}%
\begin{APACrefauthors}%
Hendrycks, D.%
, Burns, C.%
, Basart, S.%
, Critch, A.%
, Li, J.%
, Song, D.%
\BCBL {}\ \BBA {} Steinhardt, J.%
\end{APACrefauthors}%
\unskip\
\newblock
\APACrefYearMonthDay{2023}{}{}.
\newblock
\APACrefbtitle {Aligning AI With Shared Human Values.} {Aligning ai with shared human values.}
\newblock
\begin{APACrefURL} \url{https://arxiv.org/abs/2008.02275} \end{APACrefURL}
\PrintBackRefs{\CurrentBib}

\bibitem [\protect \citeauthoryear {%
Hendrycks%
\ \protect \BOthers {.}}{%
Hendrycks%
\ \protect \BOthers {.}}{%
{\protect \APACyear {2022}}%
}]{%
hendrycks2022jiminycricketdoagents}
\APACinsertmetastar {%
hendrycks2022jiminycricketdoagents}%
\begin{APACrefauthors}%
Hendrycks, D.%
, Mazeika, M.%
, Zou, A.%
, Patel, S.%
, Zhu, C.%
, Navarro, J.%
\BDBL {}Steinhardt, J.%
\end{APACrefauthors}%
\unskip\
\newblock
\APACrefYearMonthDay{2022}{}{}.
\newblock
\APACrefbtitle {What Would Jiminy Cricket Do? Towards Agents That Behave Morally.} {What would jiminy cricket do? towards agents that behave morally.}
\newblock
\begin{APACrefURL} \url{https://arxiv.org/abs/2110.13136} \end{APACrefURL}
\PrintBackRefs{\CurrentBib}

\bibitem [\protect \citeauthoryear {%
Hoffman%
, Blei%
, Wang%
\BCBL {}\ \BBA {} Paisley%
}{%
Hoffman%
\ \protect \BOthers {.}}{%
{\protect \APACyear {2013}}%
}]{%
hoffman2013stochasticvariationalinference}
\APACinsertmetastar {%
hoffman2013stochasticvariationalinference}%
\begin{APACrefauthors}%
Hoffman, M.%
, Blei, D\BPBI M.%
, Wang, C.%
\BCBL {}\ \BBA {} Paisley, J.%
\end{APACrefauthors}%
\unskip\
\newblock
\APACrefYearMonthDay{2013}{}{}.
\newblock
\APACrefbtitle {Stochastic Variational Inference.} {Stochastic variational inference.}
\newblock
\begin{APACrefURL} \url{https://arxiv.org/abs/1206.7051} \end{APACrefURL}
\PrintBackRefs{\CurrentBib}

\bibitem [\protect \citeauthoryear {%
Holland%
}{%
Holland%
}{%
{\protect \APACyear {1992}}%
}]{%
holland1992adaptation}
\APACinsertmetastar {%
holland1992adaptation}%
\begin{APACrefauthors}%
Holland, J\BPBI H.%
\end{APACrefauthors}%
\unskip\
\newblock
\APACrefYear{1992}.
\newblock
\APACrefbtitle {Adaptation in natural and artificial systems: an introductory analysis with applications to biology, control, and artificial intelligence} {Adaptation in natural and artificial systems: an introductory analysis with applications to biology, control, and artificial intelligence}.
\newblock
\APACaddressPublisher{}{MIT press}.
\PrintBackRefs{\CurrentBib}

\bibitem [\protect \citeauthoryear {%
Hong%
\ \protect \BOthers {.}}{%
Hong%
\ \protect \BOthers {.}}{%
{\protect \APACyear {2024}}%
}]{%
hong2024metagptmetaprogrammingmultiagent}
\APACinsertmetastar {%
hong2024metagptmetaprogrammingmultiagent}%
\begin{APACrefauthors}%
Hong, S.%
, Zhuge, M.%
, Chen, J.%
, Zheng, X.%
, Cheng, Y.%
, Zhang, C.%
\BDBL {}Schmidhuber, J.%
\end{APACrefauthors}%
\unskip\
\newblock
\APACrefYearMonthDay{2024}{}{}.
\newblock
\APACrefbtitle {{MetaGPT}: Meta Programming for A Multi-Agent Collaborative Framework.} {{MetaGPT}: Meta programming for a multi-agent collaborative framework.}
\newblock
\begin{APACrefURL} \url{https://arxiv.org/abs/2308.00352} \end{APACrefURL}
\PrintBackRefs{\CurrentBib}

\bibitem [\protect \citeauthoryear {%
Hume%
}{%
Hume%
}{%
{\protect \APACyear {1751}}%
}]{%
Hume1751-HUMAEC-11}
\APACinsertmetastar {%
Hume1751-HUMAEC-11}%
\begin{APACrefauthors}%
Hume, D.%
\end{APACrefauthors}%
\unskip\
\newblock
\APACrefYear{1751}.
\newblock
\APACrefbtitle {An Enquiry Concerning the Principles of Morals} {An enquiry concerning the principles of morals}.
\newblock
\APACaddressPublisher{New York}{Oxford University Press UK}.
\PrintBackRefs{\CurrentBib}

\bibitem [\protect \citeauthoryear {%
McIntyre%
, Kallada%
, Miguel%
, Feher~de Silva%
\BCBL {}\ \BBA {} Netto%
}{%
McIntyre%
\ \protect \BOthers {.}}{%
{\protect \APACyear {{\protect \bibnodate {}}}}%
}]{%
McIntyre_neat-python}
\APACinsertmetastar {%
McIntyre_neat-python}%
\begin{APACrefauthors}%
McIntyre, A.%
, Kallada, M.%
, Miguel, C\BPBI G.%
, Feher~de Silva, C.%
\BCBL {}\ \BBA {} Netto, M\BPBI L.%
\end{APACrefauthors}%
\unskip\
\newblock
\APACrefYearMonthDay{{\protect \bibnodate {}}}{}{}.
\newblock
\APACrefbtitle {{neat-python}.} {{neat-python}.}
\PrintBackRefs{\CurrentBib}

\bibitem [\protect \citeauthoryear {%
Nahian%
, Frazier%
, Harrison%
\BCBL {}\ \BBA {} Riedl%
}{%
Nahian%
\ \protect \BOthers {.}}{%
{\protect \APACyear {2021}}%
}]{%
nahian2021trainingvaluealignedreinforcementlearning}
\APACinsertmetastar {%
nahian2021trainingvaluealignedreinforcementlearning}%
\begin{APACrefauthors}%
Nahian, M\BPBI S\BPBI A.%
, Frazier, S.%
, Harrison, B.%
\BCBL {}\ \BBA {} Riedl, M.%
\end{APACrefauthors}%
\unskip\
\newblock
\APACrefYearMonthDay{2021}{}{}.
\newblock
\APACrefbtitle {Training Value-Aligned Reinforcement Learning Agents Using a Normative Prior.} {Training value-aligned reinforcement learning agents using a normative prior.}
\newblock
\begin{APACrefURL} \url{https://arxiv.org/abs/2104.09469} \end{APACrefURL}
\PrintBackRefs{\CurrentBib}

\bibitem [\protect \citeauthoryear {%
Neal%
}{%
Neal%
}{%
{\protect \APACyear {1996}}%
}]{%
bayesianlearningforneuralnetworks}
\APACinsertmetastar {%
bayesianlearningforneuralnetworks}%
\begin{APACrefauthors}%
Neal, R\BPBI M.%
\end{APACrefauthors}%
\unskip\
\newblock
\APACrefYear{1996}.
\newblock
\APACrefbtitle {Bayesian Learning for Neural Networks} {Bayesian learning for neural networks}.
\newblock
\APACaddressPublisher{Berlin, Heidelberg}{Springer-Verlag}.
\PrintBackRefs{\CurrentBib}

\bibitem [\protect \citeauthoryear {%
Neelakantan%
\ \protect \BOthers {.}}{%
Neelakantan%
\ \protect \BOthers {.}}{%
{\protect \APACyear {2022}}%
}]{%
neelakantan2022textcodeembeddingscontrastive}
\APACinsertmetastar {%
neelakantan2022textcodeembeddingscontrastive}%
\begin{APACrefauthors}%
Neelakantan, A.%
, Xu, T.%
, Puri, R.%
, Radford, A.%
, Han, J\BPBI M.%
, Tworek, J.%
\BDBL {}Weng, L.%
\end{APACrefauthors}%
\unskip\
\newblock
\APACrefYearMonthDay{2022}{}{}.
\newblock
\APACrefbtitle {Text and Code Embeddings by Contrastive Pre-Training.} {Text and code embeddings by contrastive pre-training.}
\newblock
\begin{APACrefURL} \url{https://arxiv.org/abs/2201.10005} \end{APACrefURL}
\PrintBackRefs{\CurrentBib}

\bibitem [\protect \citeauthoryear {%
Omohundro%
}{%
Omohundro%
}{%
{\protect \APACyear {2008}}%
}]{%
basicdrives}
\APACinsertmetastar {%
basicdrives}%
\begin{APACrefauthors}%
Omohundro, S\BPBI M.%
\end{APACrefauthors}%
\unskip\
\newblock
\APACrefYearMonthDay{2008}{}{}.
\newblock
{\BBOQ}\APACrefatitle {The Basic AI Drives} {The basic ai drives}.{\BBCQ}
\newblock
\BIn{} \APACrefbtitle {Proceedings of the 2008 Conference on Artificial General Intelligence 2008: Proceedings of the First AGI Conference} {Proceedings of the 2008 conference on artificial general intelligence 2008: Proceedings of the first agi conference}\ (\BPG~483–492).
\newblock
\APACaddressPublisher{NLD}{IOS Press}.
\PrintBackRefs{\CurrentBib}

\bibitem [\protect \citeauthoryear {%
Pan%
\ \protect \BOthers {.}}{%
Pan%
\ \protect \BOthers {.}}{%
{\protect \APACyear {2023}}%
}]{%
pan2023rewardsjustifymeansmeasuring}
\APACinsertmetastar {%
pan2023rewardsjustifymeansmeasuring}%
\begin{APACrefauthors}%
Pan, A.%
, Chan, J\BPBI S.%
, Zou, A.%
, Li, N.%
, Basart, S.%
, Woodside, T.%
\BDBL {}Hendrycks, D.%
\end{APACrefauthors}%
\unskip\
\newblock
\APACrefYearMonthDay{2023}{}{}.
\newblock
\APACrefbtitle {Do the Rewards Justify the Means? Measuring Trade-Offs Between Rewards and Ethical Behavior in the MACHIAVELLI Benchmark.} {Do the rewards justify the means? measuring trade-offs between rewards and ethical behavior in the machiavelli benchmark.}
\newblock
\begin{APACrefURL} \url{https://arxiv.org/abs/2304.03279} \end{APACrefURL}
\PrintBackRefs{\CurrentBib}

\bibitem [\protect \citeauthoryear {%
Park%
\ \protect \BOthers {.}}{%
Park%
\ \protect \BOthers {.}}{%
{\protect \APACyear {2023}}%
}]{%
park2023generativeagentsinteractivesimulacra}
\APACinsertmetastar {%
park2023generativeagentsinteractivesimulacra}%
\begin{APACrefauthors}%
Park, J\BPBI S.%
, O'Brien, J\BPBI C.%
, Cai, C\BPBI J.%
, Morris, M\BPBI R.%
, Liang, P.%
\BCBL {}\ \BBA {} Bernstein, M\BPBI S.%
\end{APACrefauthors}%
\unskip\
\newblock
\APACrefYearMonthDay{2023}{}{}.
\newblock
\APACrefbtitle {Generative Agents: Interactive Simulacra of Human Behavior.} {Generative agents: Interactive simulacra of human behavior.}
\newblock
\begin{APACrefURL} \url{https://arxiv.org/abs/2304.03442} \end{APACrefURL}
\PrintBackRefs{\CurrentBib}

\bibitem [\protect \citeauthoryear {%
Stanley%
\ \BBA {} Miikkulainen%
}{%
Stanley%
\ \BBA {} Miikkulainen%
}{%
{\protect \APACyear {2002}}%
}]{%
stanley2002evolving}
\APACinsertmetastar {%
stanley2002evolving}%
\begin{APACrefauthors}%
Stanley, K\BPBI O.%
\BCBT {}\ \BBA {} Miikkulainen, R.%
\end{APACrefauthors}%
\unskip\
\newblock
\APACrefYearMonthDay{2002}{}{}.
\newblock
{\BBOQ}\APACrefatitle {Evolving neural networks through augmenting topologies} {Evolving neural networks through augmenting topologies}.{\BBCQ}
\newblock
\APACjournalVolNumPages{Evolutionary computation}{10}{2}{99--127}.
\PrintBackRefs{\CurrentBib}

\bibitem [\protect \citeauthoryear {%
Vaswani%
\ \protect \BOthers {.}}{%
Vaswani%
\ \protect \BOthers {.}}{%
{\protect \APACyear {2023}}%
}]{%
vaswani2023attentionneed}
\APACinsertmetastar {%
vaswani2023attentionneed}%
\begin{APACrefauthors}%
Vaswani, A.%
, Shazeer, N.%
, Parmar, N.%
, Uszkoreit, J.%
, Jones, L.%
, Gomez, A\BPBI N.%
\BDBL {}Polosukhin, I.%
\end{APACrefauthors}%
\unskip\
\newblock
\APACrefYearMonthDay{2023}{}{}.
\newblock
\APACrefbtitle {Attention Is All You Need.} {Attention is all you need.}
\newblock
\begin{APACrefURL} \url{https://arxiv.org/abs/1706.03762} \end{APACrefURL}
\PrintBackRefs{\CurrentBib}

\bibitem [\protect \citeauthoryear {%
Wang%
, Chiu%
\BCBL {}\ \BBA {} Chiu%
}{%
Wang%
\ \protect \BOthers {.}}{%
{\protect \APACyear {2023}}%
}]{%
wang2023humanoidagentsplatformsimulating}
\APACinsertmetastar {%
wang2023humanoidagentsplatformsimulating}%
\begin{APACrefauthors}%
Wang, Z.%
, Chiu, Y\BPBI Y.%
\BCBL {}\ \BBA {} Chiu, Y\BPBI C.%
\end{APACrefauthors}%
\unskip\
\newblock
\APACrefYearMonthDay{2023}{}{}.
\newblock
\APACrefbtitle {Humanoid Agents: Platform for Simulating Human-like Generative Agents.} {Humanoid agents: Platform for simulating human-like generative agents.}
\newblock
\begin{APACrefURL} \url{https://arxiv.org/abs/2310.05418} \end{APACrefURL}
\PrintBackRefs{\CurrentBib}

\bibitem [\protect \citeauthoryear {%
Yang%
\ \protect \BOthers {.}}{%
Yang%
\ \protect \BOthers {.}}{%
{\protect \APACyear {2024}}%
}]{%
yang2024psychogatnovelpsychologicalmeasurement}
\APACinsertmetastar {%
yang2024psychogatnovelpsychologicalmeasurement}%
\begin{APACrefauthors}%
Yang, Q.%
, Wang, Z.%
, Chen, H.%
, Wang, S.%
, Pu, Y.%
, Gao, X.%
\BDBL {}Huang, G.%
\end{APACrefauthors}%
\unskip\
\newblock
\APACrefYearMonthDay{2024}{}{}.
\newblock
\APACrefbtitle {PsychoGAT: A Novel Psychological Measurement Paradigm through Interactive Fiction Games with LLM Agents.} {Psychogat: A novel psychological measurement paradigm through interactive fiction games with llm agents.}
\newblock
\begin{APACrefURL} \url{https://arxiv.org/abs/2402.12326} \end{APACrefURL}
\PrintBackRefs{\CurrentBib}

\end{thebibliography}

\newpage
\appendix

\clearpage

\end{document}